\documentclass{article}
\usepackage{spconf,amsmath,xcolor,graphicx}
\usepackage{multirow,booktabs}
\usepackage{url}
\usepackage{hyperref}
\usepackage{stfloats}
\usepackage{subcaption}
\usepackage{amssymb}
\usepackage[absolute,overlay]{textpos}

\title{Is Perturbation-based Image Protection Disruptive to Image Editing?}
\name{
Qiuyu Tang\textsuperscript{1}, Bonor Ayambem\textsuperscript{1}, Mooi Choo Chuah\textsuperscript{1}, Aparna Bharati\textsuperscript{1}
}
\address{\textsuperscript{1} Department of Computer Science and Engineering, Lehigh University, Bethlehem, PA, USA.}

\begin{document}

\begin{textblock}{195}(10,10) 
  {\small Copyright 2025 IEEE. Published in 2025 IEEE International Conference on Image Processing (ICIP), scheduled for 14-17 September 2025 in Anchorage, Alaska, USA. Personal use of this material is permitted. However, permission to reprint/republish this material for advertising or promotional purposes or for creating new collective works for resale or redistribution to servers or lists, or to reuse any copyrighted component of this work in other works, must be obtained from the IEEE. Contact: Manager, Copyrights and Permissions / IEEE Service Center / 445 Hoes Lane / P.O. Box 1331 / Piscataway, NJ 08855-1331, USA. Telephone: + Intl. 908-562-3966.}
\end{textblock}

\maketitle

\begin{abstract}
The remarkable image generation capabilities of state-of-the-art diffusion models, such as Stable Diffusion,
can also be misused 
to spread misinformation and plagiarize copyrighted materials. To mitigate the potential risks associated with image editing, current image protection methods rely on adding imperceptible perturbations to images to obstruct diffusion-based editing. 
A fully successful protection for an image implies that the output of editing attempts is an undesirable, noisy image which is completely unrelated to the reference image. 
In our experiments with various perturbation-based image protection methods across multiple domains (natural scene images and artworks) and editing tasks (image-to-image generation and style editing), we discover that such protection does not achieve this goal completely. In most scenarios, diffusion-based editing of protected images generates a desirable output image which adheres precisely to the guidance prompt.
Our findings suggest that adding noise to images may paradoxically increase their association with given text prompts during the generation process, leading to unintended consequences such as \textit{better} resultant edits. 
Hence, we argue that perturbation-based methods may not provide a sufficient solution for robust image protection against diffusion-based editing. 
\footnote{Supplemental Material (SM): \footnotesize{\url{https://dx.doi.org/10.60864/8fp1-q828}}; Code/data: {\footnotesize\url{https://github.com/CV-Lehigh/perturb2protect-disruptive-or-helpful}}.}
\end{abstract}


\vspace{-10pt}
\section{Introduction}
\label{sec:introduction}
\vspace{-5pt}
Generative models~\cite{goodfellow2014generative, karras2019style, rombach2022high, ruiz2023dreambooth, gal2022image, kumari2023multi} have become increasingly popular and have been enthusiastically adopted within the AI research and creative arts communities. Among various generative techniques, diffusion-based models~\cite{rombach2022high, ruiz2023dreambooth, gal2022image, kumari2023multi, meng2021sdedit} are widely favored for their high-quality image generation and straightforward usability. However, their adoption has had a dichotomized effect. On one hand, the ability to generate and edit hyperrealistic images with minimal input from the user has enabled any layperson with almost no artistic skills to undertake enhanced creative pursuits. On the other hand, this ability has also raised concerns regarding misinformation, plagiarism and copyright violation~\cite{nyt_ai_art_2022, nyt_ai_generated_art_2022}. 
\begin{figure}[t]
    \centering
    \includegraphics[width=0.9\linewidth]{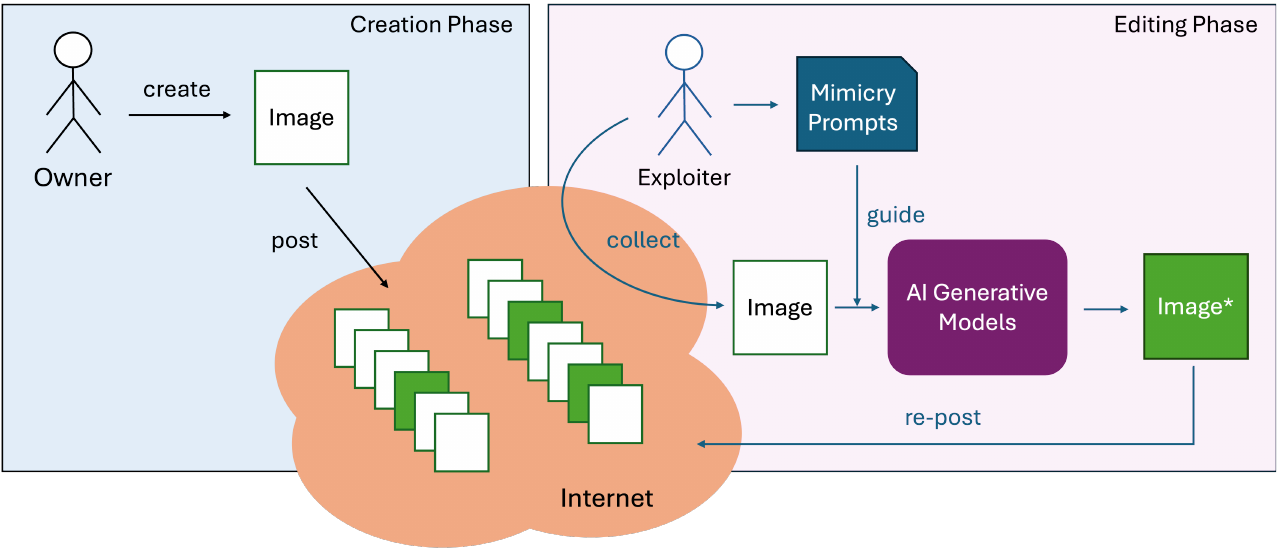}
    \vspace{-10pt}
    \caption{Scenario of Malicious Exploitation: The \textit{Owner} creates an image and posts it online. The \textit{Exploiter} collects the original image from the Internet, feeds it into a diffusion-based model with a mimicry prompt as guidance to generate a similar but manipulated image, and then reposts the counterfeit image back to the Internet. 
    }
    \vspace{-20pt}
    \label{fig:scenario}
\end{figure}

Fig.~\ref{fig:scenario} demonstrates a potential exploitative scenario using diffusion-based generative models. 
Such exploitation can typically follow two approaches: inference-based editing~\cite{rombach2022high} and fine-tuning-based editing~\cite{ruiz2023dreambooth, gal2022image}. Inference-based editing involves the direct use of pre-trained models, where exploiters craft prompts to manipulate images without modifying the model's parameters. In contrast, fine-tuning-based editing methods
adapt models by retraining them on specific data, enabling exploiters to produce highly customized content, such as replicating specific styles or generating images resembling particular individuals. While fine-tuned models produce higher-quality results, they are computationally expensive and challenging for non-experts. Hence, this paper focuses on more accessible inference-based editing techniques and methods to protect against them.

Specifically, this paper delves into the effectiveness of current protection methods against inference-based editing, which mostly employ adversarial attacks on diffusion-based models by injecting imperceptible perturbations designed to disrupt the model's performance. The output of the model relies on two inputs, the original/reference image and the text prompt guiding the edit. As long as the output image shares some content with the reference and the generated edits align with the guidance prompts, malicious exploitation is considered successful, i.e., the exploiter achieves the desired output. 
Therefore, our experiments across diverse tasks and datasets focus on analyzing how closely the content generated from protected and unprotected image variants aligns with the prompt, providing insight into the robustness of protection methods. 
The results using multiple measures~\cite{hessel2021clipscore, sarto2023positive, sarto2024positive} show that 
perturbations as a byproduct make the edited image generated from the protected image align more with text prompts while sharing content with the original images, leading to preferable results for exploiters. 
In summary, our contributions are:
\vspace{-5pt}
\begin{enumerate}
\itemsep-0.1em
    \item \textbf{Evaluation of Protection Methods}: Investigating the effectiveness of current protection methods against inference-based editing by analyzing the alignment between generated content and guidance prompts, providing insights into their usefulness and robustness.

    \item \textbf{Comprehensive Domain Analysis}: Experiments considering natural scene and artwork images for potential exploitation through content and style-based editing.

    \item \textbf{Theoretical Explanation}: We provide a theoretical explanation for the above observations.
    
\end{enumerate}

\vspace{-15pt}
\section{Related Work}
\label{sec:relate}
\vspace{-5pt}

\noindent\textbf{Diffusion-based Models} -
\label{sec: diffusion}
Models such as~\cite{rombach2022high, meng2021sdedit} have demonstrated superior image generation performance by faithfully adhering to user input and ensuring realism in the generated images.
They achieve this via a forward process where noise is added to an image and a reverse process which denoises them.
Stable Diffusion (SD)~\cite{rombach2022high}, a popular model, implements the Latent Diffusion Model (LDM), which operates in latent space rather than pixel space, significantly reducing computational time and cost.
SD supports applications such as Inpainting, Text-to-Image, and Image-to-Image editing using conditional inputs like text prompts, segmentation maps, and reference images.
For our context of protecting image data, we focus on Image-to-Image generation, which takes an input image and text prompt to generate an image. 
These models also allow personalization~\cite{ruiz2023dreambooth, gal2022image, kumari2023multi}, where pre-trained models can be fine-tuned using a small set of images to enable the learning of new concepts.
This allows users to generate images of specific subjects, such as objects, art styles, or identities. 
Note that SD~\cite{rombach2022high} versions (1.x), pre-trained on millions of images from the LAION-5B dataset~\cite{schuhmann2022laion},
typically consists of approximately 860 million parameters. Consequently, fine-tuning is computationally expensive and requires adaptations to make it feasible~\cite{ruiz2023dreambooth, gal2022image, kumari2023multi}. While customization yields high-quality mimicked output, 
direct image editing using a pre-trained Stable Diffusion model~\cite{rombach2022high} is accessible to exploiters with minimal coding knowledge and limited computational resources. 
Therefore, we identify inference-based editing as a greater risk and evaluate protection methods against it.

\noindent\textbf{Protection Methods} -
\label{sec: protection}
In mimicry scenario (see Fig.~\ref{fig:scenario}), images
serve as the input that is manipulated, models provide the mechanism for generating edits, and mimicry prompts guide the generation toward the desired manipulated output. Since the process relies entirely on the interaction of these components, addressing them directly is the only viable option for developing effective countermeasures. 
There can be limited reliance on prompts because it is difficult to anticipate and enumerate all the prompts that exploiter might use. New generative models~\cite{schramowski2023safe} that detect and counteract inappropriate generation aren't sufficient, as exploiters retain the flexibility to choose models at their discretion. Therefore, numerous protection measures~\cite{salman2023raising, shan2023glaze, van2023anti, liu2024metacloak, tang2025watermarks} operate on input images and transform clean images 
by adding adversarial perturbations that distort the behavior of diffusion-based models. 

Among protection methods against inference-based editing~\cite{salman2023raising, shan2023glaze, liang2023mist}, PhotoGuard~\cite{salman2023raising} is the first to utilize adversarial perturbations. When a protected image is attempted to be edited, the goal is to generate an obviously manipulated image, i.e., unrealistic and unrelated to the original. The authors propose two attacks towards SD~\cite{rombach2022high}: Encoder-Attack shifts the representation of the input image to a target latent representation, whereas Diffusion-Attack shifts the output image towards a target image (e.g., a black image). 

Liang et al.~\cite{liang2023mist, liang2023adversarial} and Glaze~\cite{shan2023glaze} investigate the risks of generative models being exploited to create counterfeit artworks.
Their goal is to prevent generative models from effectively extracting features to imitate style or content. AdvDM~\cite{liang2023adversarial} introduces a semantic loss function for generating adversarial examples via Monte Carlo estimation, while Mist~\cite{liang2023mist} extends this loss by combining it with textual loss from PhotoGuard~\cite{salman2023raising}. Glaze~\cite{shan2023glaze}, a black-box tool, applies style cloaks to shift an artwork's representation in the generator's feature space toward unrelated styles.
These protection methods share the underlying approach of introducing adversarial perturbations within a small budget on clean images, which degrades diffusion-based generative models by producing low-quality or unrelated edited images.

Prior study~\cite{honig2024adversarial} focuses on post-hoc defenses, while our work takes a complementary approach by evaluating whether perturbations can proactively disrupt editing success, shifting the emphasis from recovery to prevention.

\vspace{-10pt}
\section{Impact of Protection on Image Editing}
\label{experiments}
\vspace{-5pt}
To evaluate the performance of image protection against inference-based editing via pre-trained Stable Diffusion~\cite{rombach2022high}, we conduct experiments on three state-of-the-art protection methods that employ meticulously crafted adversarial perturbations, PhotoGuard \cite{salman2023raising}, Mist~\cite{liang2023mist}, and Glaze~\cite{shan2023glaze}. 
PhotoGuard protects natural scene images, while Mist and Glaze protect artwork images. 
Considering potential exploitation scenarios and editing applications, we conduct experiments across different domains, including natural scenes and artistic images.
We measure the effectiveness of protection methods by evaluating whether the exploiter gets a realistic and related output image when editing a protected image. When the generated images from protected samples look real and closely adhere to the text prompts, exploiters achieve their intended outcome, indicating that the protection is not fully effective. 



\noindent\textbf{Generation Configurations.}
We use Stable Diffusion (v1.5)~\cite{rombach2022high} as the pre-trained image generator for image editing in our experiments. To ensure the non-randomness and reproducibility of the results, we choose five distinct seeds for the generator: 9222, 999, 123, 66, and 42.
Other hyperparameters, including guidance scale, strength, and total steps in the generation, are maintained at their default settings as in~\cite{salman2023raising}. 

\begin{figure}[t]
    \centering
    \includegraphics[width=0.9\linewidth]{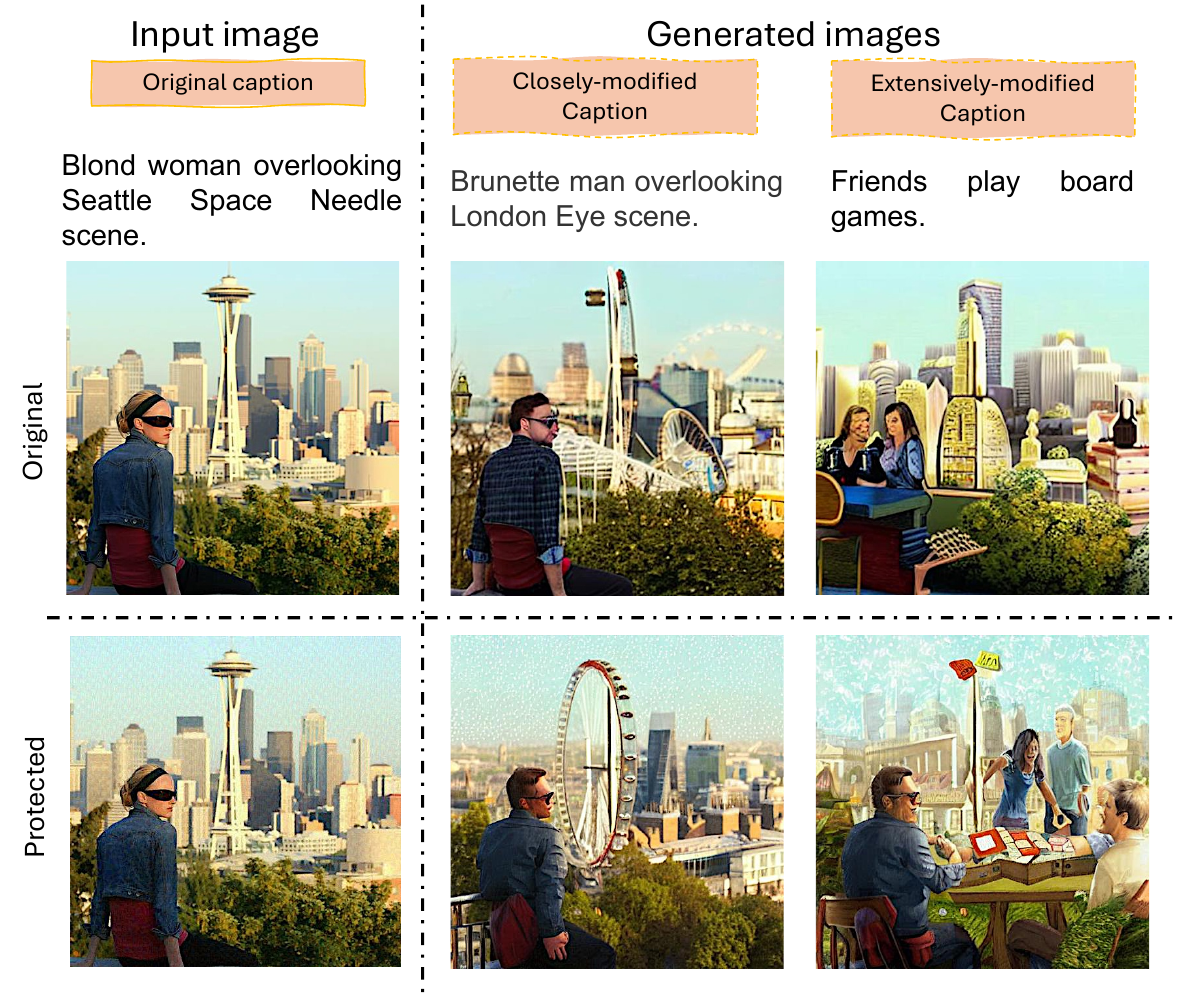}
    \vspace{-15pt}
    \caption{Image-to-Image Generation results from Stable Diffusion v1.5.~\cite{rombach2022high} for images protected by PhotoGuard~\cite{salman2023raising}, using captions modified with small and large semantic changes.}
    \label{fig:visual1}
    \vspace{-15pt}
\end{figure}

\vspace{-10pt}
\subsection{Image-to-Image Generation}
\vspace{-5pt}
To evaluate protection performance on natural scene images, we implement PhotoGuard~\cite{salman2023raising} on the Flickr8k~\cite{flickr8k} dataset because we can use modified captions as text prompts to obtain edits. Flickr8k~\cite{flickr8k} dataset contains over 8,000 general images, each paired with up to five detailed captions.

We generate two sets of modified captions from the first caption for each image with the assistance of Claude's Sonnet 3.5~\cite{claude2024}. One set with prompts that are contextually \textit{close} to the original captions, and another with prompts that are contextually \textit{far}. For example, given the original caption: \textit{A young girl in a pink dress going into a wooden cabin}, a \textit{close} prompt could be: \textit{A young boy in a blue shirt going into a brick house}; whereas a \textit{far} prompt could be: \textit{Two cats lounging on a couch}. Usually, a \textit{close} prompt can be constructed by substituting a few nouns and adjectives from the original caption with semantically similar ones. The \textit{far} captions on the other hand, were obtained by prompting the LLM to create prompts that are contextually very different from the original captions. The generated captions were manually verified for quality and semantic relevance. 
To quantitatively evaluate the semantic similarity between the original captions and the modified prompts, we employ the Universal Sentence Encoder~\cite{cer2018universal}. 
Additional examples of prompts with corresponding semantic similarity scores are provided in SM Sec.1. 

Each image and its protected variant are edited using the close and the far guidance prompts from the two sets. Fig.~\ref{fig:visual1} visualizes some inputs and their generation outputs.
The average Blind/Referenceless Image Spatial Quality Evaluator (BRISQUE)~\cite{mittal2012no} score for the generated images corresponding to two captions is 17.88 (close caption: 17.82, far caption: 17.94), compared to 22.27 for the original images. Thus, the generated images exhibit comparable quality to the original. 

\begin{figure}[t]
    \centering
    \includegraphics[width=\linewidth]{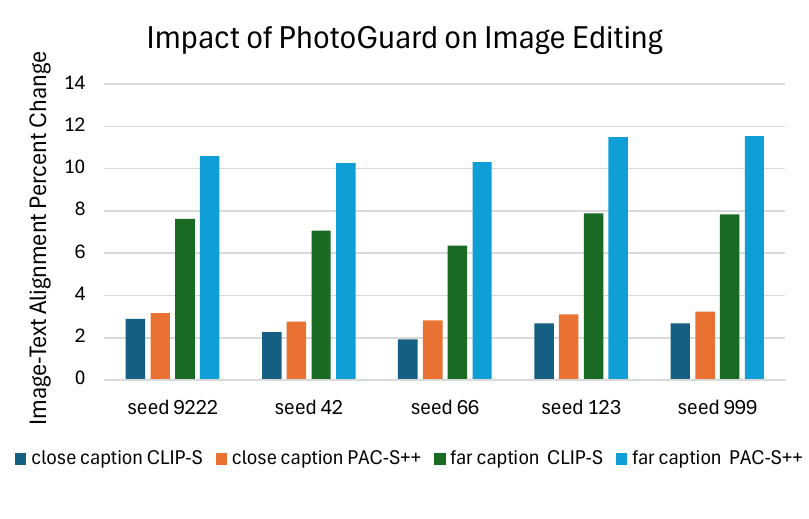}
    \vspace{-35pt}
    \caption{The impact of protection on Flickr8k~\cite{flickr8k} dataset over 5 seeds and close/far captions. Image-Text Alignment is measured by CLIP-S~\cite{hessel2021clipscore} and PAC-S++~\cite{sarto2024positive}. }
    \vspace{-15pt}
    \label{fig:impactI2I}
\end{figure}

\noindent\textbf{Metrics.}
To assess the effectiveness of protection perturbations, we capture how well the generated images align with the text prompts. Therefore, we consider semantic similarity metrics that compare the association of visual-textual content~\cite{krinsky2024exploring}. CLIP-S~\cite{hessel2021clipscore} is a modified cosine similarity measure between the image and candidate caption representations generated using the large multimodal model CLIP~\cite{radford2021learning}. PAC-S++~\cite{sarto2023positive, sarto2024positive} is another score that augments the positive samples by generative models and achieves a captioning metric that is more accurate and consistent with human evaluation. 
Such Image-Text Alignment metrics (ITA Score) quantify the extent 
of alignment between an edited image and its intended textual description, thereby indicating the degree to which an exploiter can achieve their desired output.
We define the Percentage Change (see Eq.~\ref{eq: change}) as a relative measure to evaluate the impact of adversarial protections. 
The Actual Change represents the difference in the ITA scores between the generated image from the protected image ($I_a$) and the generated image from the original clean image ($I_n$). 
By normalizing the Actual Change by the Reference, ITA for $I_n$, the Percentage Change provides a standardized measure of how the perturbation affects the model’s adherence to prompts and its effectiveness in meeting protection goals.

\vspace{-15pt}
\begin{equation}
\label{eq: change}
\resizebox{\linewidth}{!}{$
    \begin{aligned}
        \text{Percentage Change} &= \frac{\text{Actual Change in ITA}}{\text{Reference ITA}}, \\
        \text{Actual Change} &= \text{ITAScore}(I_a, \text{text}) - \text{ITAScore}(I_n, \text{text}), \\
        \text{Reference} &= \text{ITAScore}(I_n, \text{text}).
    \end{aligned}
$}
\end{equation}
\vspace{-15pt}


In Fig.~\ref{fig:impactI2I}, we compare the Image-Text Alignment (ITA) between edits from original and protected images, showing the ITA Percentage Change across five seeds for prompts that are semantically \textit{close} or \textit{far} from the original captions.

\vspace{-10pt}
\subsection{Image Stylization} 
\vspace{-5pt}
Another common malicious editing scenario is Style Mimicry, where exploiters attempt to 
edit natural scene images or existing artworks to match target famous art styles.
Stylization 
tasks are specialized because style captures the essence of artistic expression, such as color, texture, brushstrokes, and patterns, without altering the core structure or content of an image. 
The success of protection in such editing scenarios is measured by how close the content of the generated output is to the reference image and how well the style matches to the target style in the prompt.

\begin{figure}[t]
    \centering
    \includegraphics[width=\linewidth]{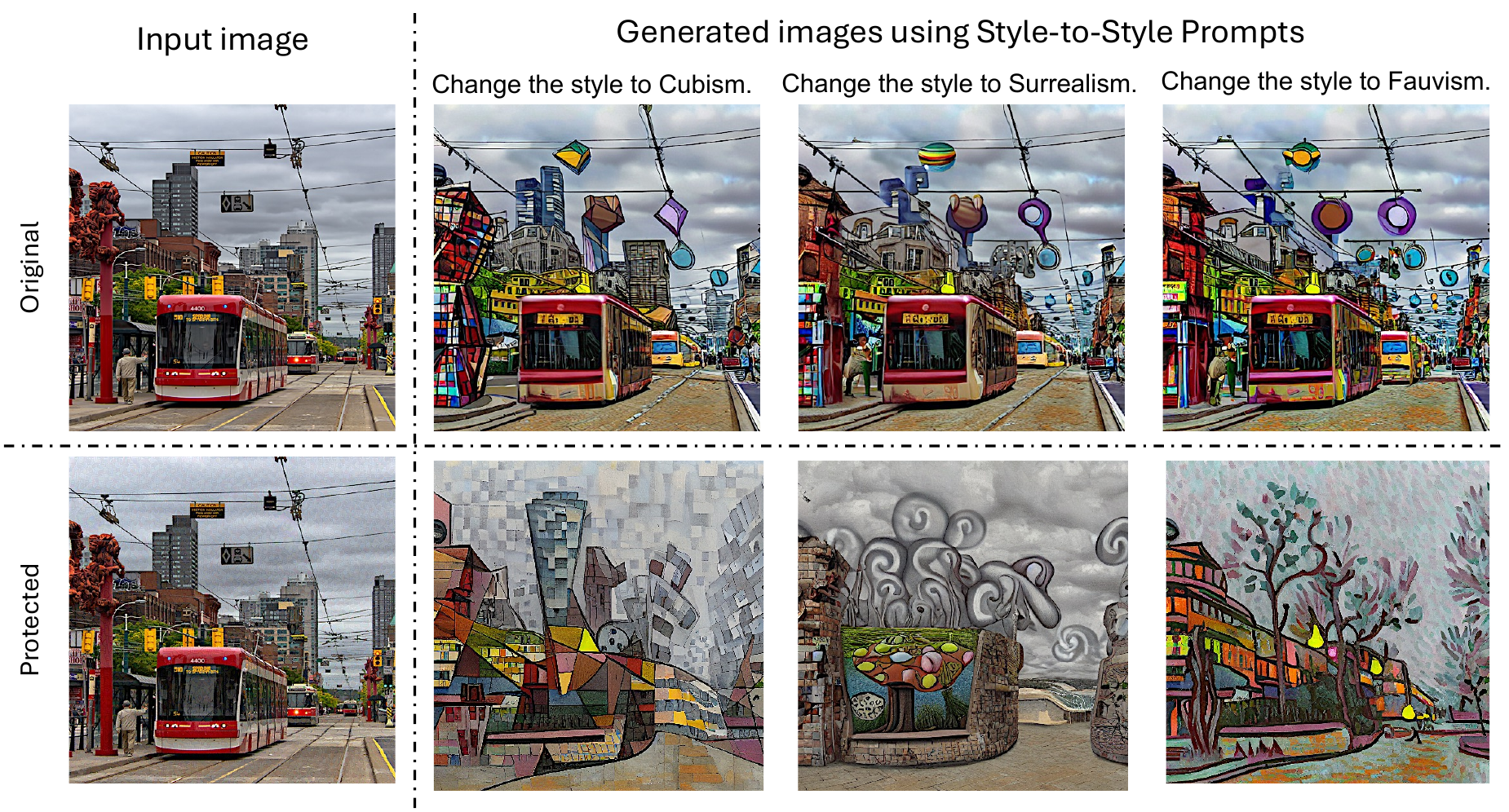}
    \vspace{-25pt}
    \caption{An original natural scene image and its protected version stylized to Cubism, Surrealism, and Fauvism.}
    \vspace{-15pt}
    \label{fig:S2Svisual}
\end{figure}

\noindent\textbf{Natural Scene Image Domain} - 
For this task, we use the Flickr1024~\cite{Flickr1024} dataset which includes more than 1,000 very high quality images.
The prompt format used during editing is ``change the style to [V]", where
[V] represents 7 famous art styles -- Cubism, Post-Impressionism, Impressionism, Surrealism, Baroque, Fauvism, and Renaissance. Fig.~\ref{fig:S2Svisual} provides a visual sequence, including an original image from Flickr1024~\cite{Flickr1024}, its adversarially protected counterpart generated using PhotoGuard~\cite{salman2023raising}, and the different stylized versions created in response to various style transfer prompts. 

\begin{figure}[h!]
    \centering
    \vspace{-10pt}
    \includegraphics[width=0.9\linewidth]{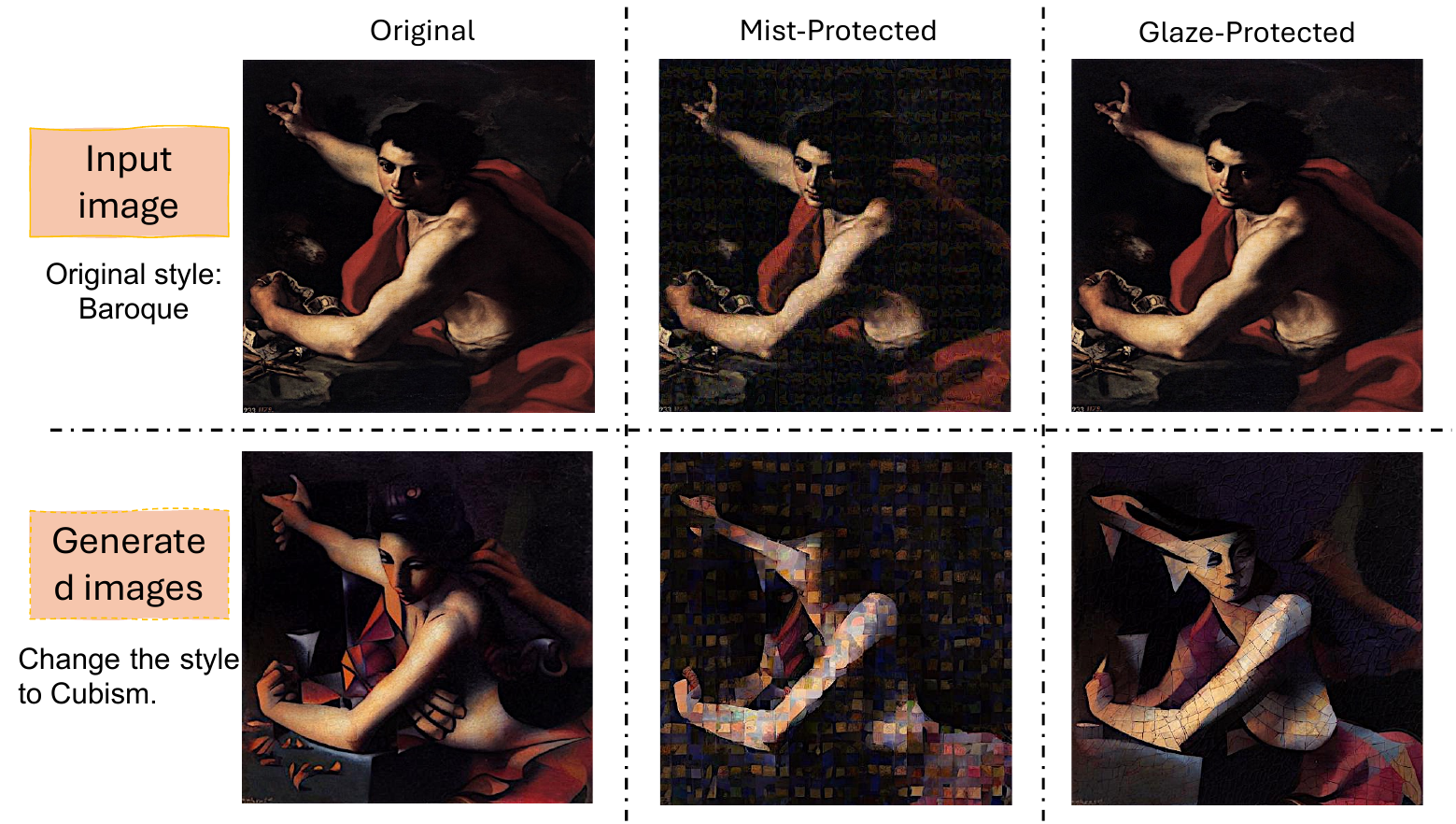}
    \vspace{-10pt}
    \caption{Qualitative image protection performance for artwork. 
    }
    \vspace{-10pt}
    \label{fig:art_visual}
\end{figure}

\noindent\textbf{Artwork Image Domain} - 
To evaluate perturbation-based art protection methods such as Glaze~\cite{shan2023glaze} and Mist~\cite{liang2023mist}, we perform style transfer on WikiArt~\cite{artgan2018} artworks.
The prompt format is the same as the natural scene images, and [V] indicates a random, unrelated style from the WikiArt~\cite{artgan2018} labels. Fig.~\ref{fig:art_visual} illustrates the input images in Baroque style and their generated variants after being transferred to an unrelated style.



\vspace{-10pt}
\section{Results \& Discussion}
\vspace{-5pt}
Our analysis for the image-to-image editing task includes four experimental settings based on the ITA scores for close and far captions as shown in Fig.~\ref{fig:impactI2I}.
The positive Percentage Changes indicate that protected images achieve stronger alignment with text prompts than original images, highlighting the influence of adversarial protections. 
Notably, all settings yield positive Percentage Changes, reinforcing the observation that 
edits derived from protected images retain more information from the text prompts.


\begin{table}[h!]
\small
    \centering
    \vspace{-5pt}
    \resizebox{0.8\linewidth}{!}{
        \begin{tabular}{l c c}
            \toprule
            \textbf{ITAScore}  & \textbf{Close caption} & \textbf{Far caption} \\ 
            \midrule
            CLIP-S~\cite{hessel2021clipscore}    & 54.23\% & 61.54\%\\
            PAC-S++~\cite{sarto2024positive}  & 61.82\% & 67.79\%\\
            \bottomrule
        \end{tabular}
    }
    \vspace{-5pt}
    \caption{Percentage of Flickr8k~\cite{flickr8k} images for which Actual Change in ITA was $\geq$ 0 using close and far captions.}
    \vspace{-5pt}
    \label{tab:tableI2I}
\end{table}

Further, we analyze the frequency of cases where the Actual Change is greater than 0. Table~\ref{tab:tableI2I} shows that the Actual Change is positive for majority cases, indicating that images with adversarial perturbations exhibit stronger alignment with the text prompts compared to their unprotected counterparts. 

\begin{figure}[t]
    \centering
    \includegraphics[width=.87\linewidth]{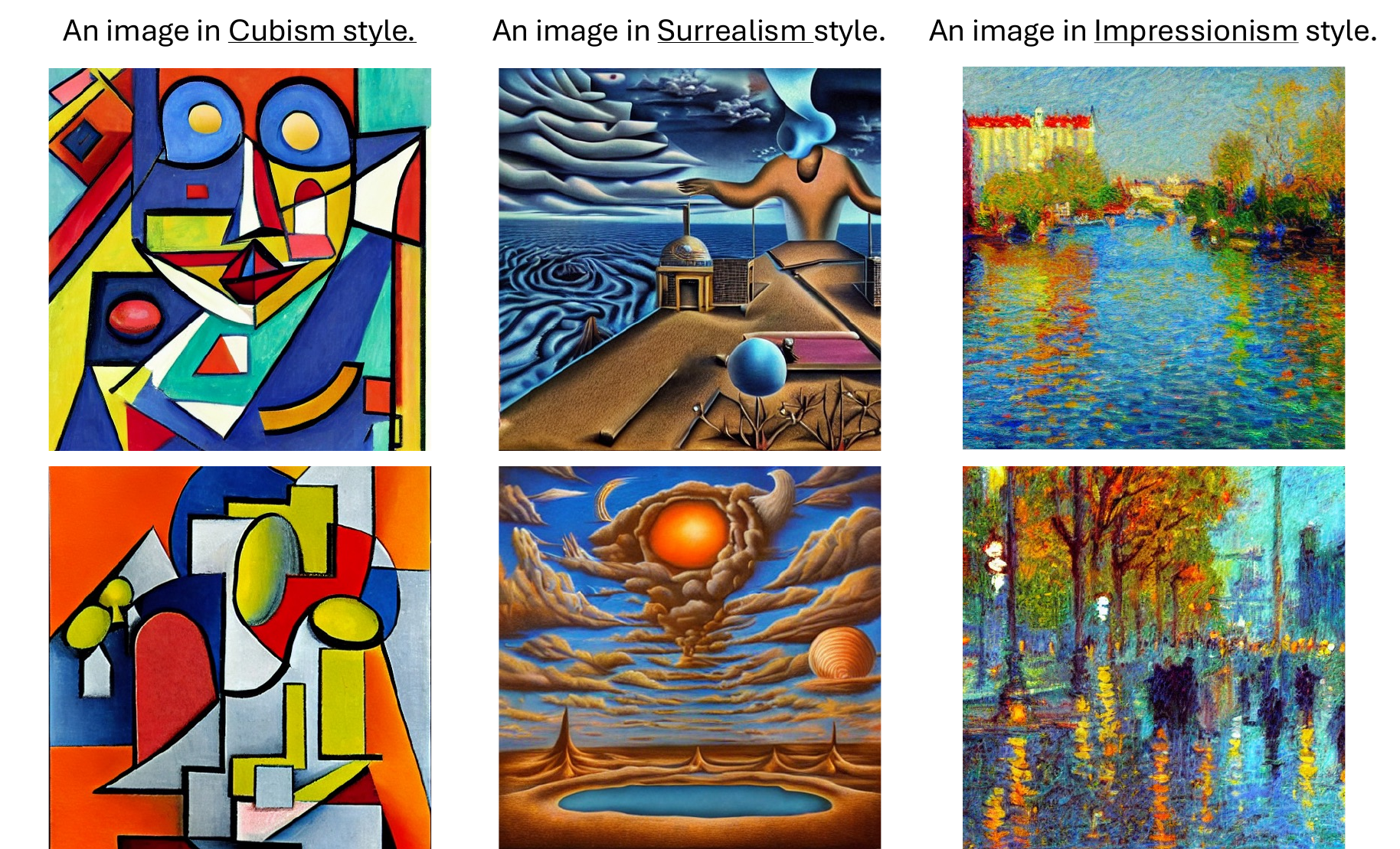}
    \vspace{-10pt}
    \caption{Text-to-image generation for three artistic styles using SD~\cite{rombach2022high} which uses CLIP~\cite{radford2021learning} encoder. }
    \vspace{-15pt}
    \label{fig:clip_capture_art}
\end{figure}

We evaluate the effectiveness of protection on stylization tasks by text-image alignment and style matching. To confirm if CLIP-S~\cite{hessel2021clipscore} and PAC-S++~\cite{sarto2024positive} understand style both in text and image space, we use SD that uses CLIP~\cite{radford2021learning} to generate images solely based on text prompts. Fig.~\ref{fig:clip_capture_art} shows that the model understands style and the ITA scores~\cite{hessel2021clipscore, sarto2024positive} can capture the semantic meaning of style. Percentage Change in ITA scores is provided in SM Sec. 2.

The stylization results follow similar trends. Visual comparisons (see Fig.~\ref{fig:S2Svisual}) indicate that generations from the protected images exhibit better alignment to the target style compared to those generated from the original images. This can be seen in fragmented and geometric forms for Picasso's Cubism, fantastical scenes portrayed in the Surrealist style and bold brushstrokes, non-naturalistic colors, and flat compositions with minimal detail for Fauvism. 
Quantitative results in Table~\ref{tab:tableS2S1} demonstrate that adversarial perturbations enhance alignment, particularly when transferring art styles to realistic images, where the effect of stylization is more pronounced.

\begin{table}[h!]
\small
    \centering
    \resizebox{\linewidth}{!}{
        \begin{tabular}{l l c c}
            \toprule
            \textbf{Protection Method} & \textbf{Dataset}  & \textbf{ITAScore} & \textbf{Actual Change $\geq$ 0} \\ 
            \midrule
            \multirow{2}{*}{PhotoGuard~\cite{salman2023raising}} & \multirow{2}{*}{Flickr1024} & CLIP-S  & \textbf{77.54\%} \\
                                               & & PAC-S++   & \textbf{68.43\%} \\
            
            \midrule
            \multirow{2}{*}{Mist~\cite{liang2023mist}} & \multirow{2}{*}{WikiArt} & CLIP-S  & \textbf{54.15\%} \\
                                          & & PAC-S++  & \textbf{56.90\%} \\
            \midrule
            \multirow{2}{*}{Glaze~\cite{shan2023glaze}} & \multirow{2}{*}{WikiArt} & CLIP-S   & \textbf{54.12\%} \\
                                           & & PAC-S++   & \textbf{56.90\%} \\
            \bottomrule
        \end{tabular}
    }
    \vspace{-10pt}
    \caption{
    Actual Change in ITA Scores for Image Stylization.
    Full results are shown in SM Sec. 3.}
    \vspace{-15pt}
    \label{tab:tableS2S1}
\end{table}

The results highlight a significant limitation of adversarial perturbations for protection. Instead of impeding alignment, adversarial perturbations often enhance the generative model's responsiveness to prompts, inadvertently enabling exploiters to produce outputs that align more closely with their objectives. Such protection is not disruptive to the image editing process and may not be able to prevent malicious agents from copying unauthorized material. The unintended consequences of using adversarial perturbations reveal vulnerabilities in existing methods and underscore the urgent need for more effective protection techniques~\cite{honig2024adversarial}.

\noindent\textbf{Theoretical Explanation} - 
\label{sec: theory}
The diffusion-based editing process typically involves an autoencoder to transition between latent and pixel-based image spaces, a U-Net to predict noise removal, and a CLIP encoder~\cite{radford2021learning} to process the text prompt.
Given that the image $x_0$ follows a real data distribution, $x_0 \sim q(x_0)$, the encoder of autoencoder $\mathcal{E}_\phi$ first embeds $x_0$ into latent variable, $z_0 = \mathcal{E}_\phi(x_0)$, parameterized by $\phi$. The forward process keeps accumulating the noise during $T$ steps, following a Markov process (see Eq.~\ref{eq: forward}). Note that $\beta$ schedules a sequence of noise values that determine how much noise is added at each step. The noise scheduler algorithm defines how many diffusion steps are needed during inference as well as how to calculate a less noisy latent. Finally, $z_T$ becomes approximately an isotropic Gaussian random variable when $\overline{\alpha}_t \to 0$ (see Eq.~\ref{eq: final forward}).
\vspace{-5pt}
\begin{equation}
\label{eq: forward}
    q(z_t|z_{t-1})=\mathcal{N}(z_t;\sqrt{1-\beta _t}z_{t-1}, \beta_t I)
\end{equation}
\begin{equation}
\label{eq: final forward}
    q(z_t|z_0)= \mathcal{N}(z_t; \sqrt{\overline{\alpha}_t}z_{t-1}, (1-\overline{\alpha}_t)I)
\end{equation}

Here, $\alpha _t = 1-\beta _t$, $\overline{\alpha}_t=\prod_{s=1}^t \alpha_s$.
Meanwhile, the LDM allows a prompt $c$ as the guidance for the generation. The prompt $c$, embedded by the CLIP~\cite{radford2021learning} text encoder, is introduced into the cross-attention layers of the U-Net, achieving the mapping from text to images. The matching process is formulated by:
\vspace{-5pt}
\begin{equation}
    \mathcal{L}_{LDM}(\theta, z_0) = \mathbb{E}_{z_0, t, \epsilon \sim \mathcal{N}(0,1),c} \| \epsilon-\epsilon_\theta (z_t,t,c)\|^2_2,
    \vspace{-5pt}
\label{eq: loss}
\end{equation}
where $\epsilon_\theta$ is the denoising network, trained by minimizing the objective function in Eq.~\ref{eq: loss}.

The distinctive aspect of using perturbation-based protection against diffusion models lies in the fact that both the target model and the attacking method leverage noise as a fundamental component. 
The diffusion mechanism in LDMs uses two Markov chains — a forward process that adds Gaussian noise and a reverse process that gradually removes it according to the U-Net’s predictions. The degree of noise removal depends partly on the text prompt’s guidance, so areas with more noise are replaced with new content while non-noisy regions remain unchanged. Therefore, text prompts guide each step of the reverse process to generate new content.

\begin{figure}[t]
    \centering
    \includegraphics[width=\linewidth]{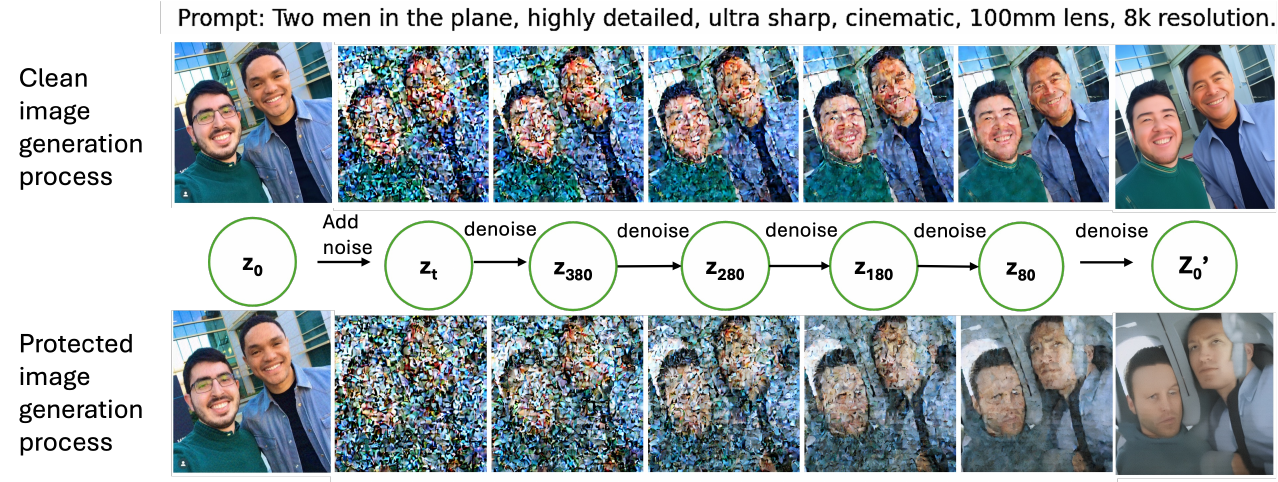}
    \vspace{-20pt}
    \caption{Comparison of the generation from clean image and PhotoGuard-protected image. For visualization, we decode interim state latents back to pixel space.}
    \vspace{-15pt}
    \label{fig:theory_visualization}
\end{figure}


Most protection methods introduce small perturbations that, though subtle, accumulate during Stable Diffusion’s generation, causing noticeable changes in the final output. As shown in Fig.~\ref{fig:theory_visualization}, with the same prompt, the protected image’s latents are noisier at each timestep, allowing for greater editing flexibility and stronger text prompt influence. Consequently, the protected image aligns more closely with the prompt (e.g., the word “plane”).

\begin{figure}[h!]
    \centering
    \vspace{-10pt}
    \includegraphics[width=\linewidth]{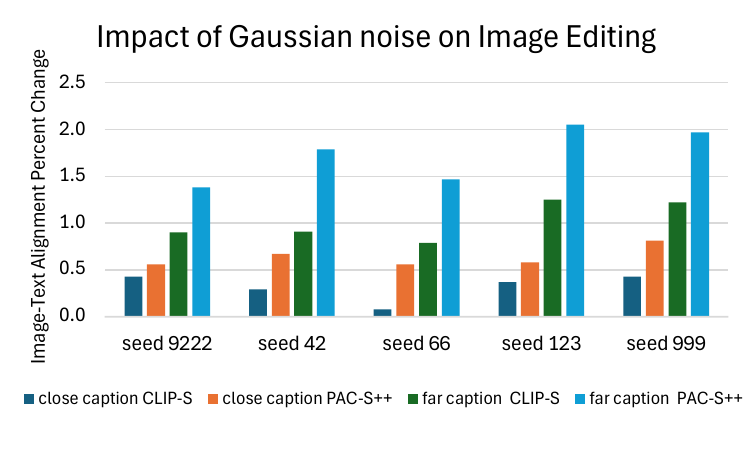}
    \vspace{-30pt}
    \caption{Results showing the impact of simulated protection with Gaussian noise on the Flickr8k~\cite{flickr8k} dataset. }
    \vspace{-15pt}
    \label{fig:impact_gaus}
\end{figure}

\noindent\textbf{Comparison with Gaussian Noise - } 
We conducted an experiment where the carefully designed perturbations~\cite{salman2023raising} were replaced with random Gaussian noise, $\mathcal{N}(0, 5^2)$.
Percentage Change in ITA for image-to-image generation with Gaussian noise as protection is presented in Fig.~\ref{fig:impact_gaus}.
The results exhibit the same trend observed in Fig.~\ref{fig:impactI2I}, with all configurations yielding positive Percentage Changes. This demonstrates that even random, unstructured noise can enhance alignment between the generated image and the prompt, confirming our theoretical understanding that the phenomenon is caused by the role of noise in protection and diffusion.


\vspace{-10pt}
\section{Conclusion}
\label{sec: conclusion}
\vspace{-5pt}
Our experiments show that perturbations can inadvertently enhance prompt alignment instead of fully disrupting image editing. 
Due to the inherent properties of diffusion-based models, these perturbations introduce additional flexibility when editing noise pixels, further reinforcing adherence to prompts. Consequently, current protection methods do not fully achieve their intended goal, underscoring the need for future solutions to explicitly evaluate and ensure disruption. We therefore advise deploying adversarial perturbations only after rigorously verifying their robustness; otherwise, malicious actors may still realize their objectives.

{
    \small
    \bibliographystyle{IEEEbib}
    \bibliography{refs_short}
}
\end{document}